\documentclass[runningheads,10pt]{llncs}

\usepackage{booktabs} 
\usepackage{amsfonts}
\usepackage{amsmath}
\usepackage{graphicx}
\usepackage{color}
\usepackage[hidelinks]{hyperref}
\usepackage{url}
\usepackage[lined,noend,algo2e,ruled,linesnumbered]{algorithm2e}
\usepackage[sort]{cite}
\usepackage{listings}
\usepackage{multirow}

\usepackage{times}
\usepackage{helvet}
\usepackage{courier}

\begin{document}

\title{On The Stability of Interpretable Models}

\author{
    Riccardo Guidotti \and
    Salvatore Ruggieri
}
\authorrunning{R. Guidotti and S. Ruggieri}

\institute{
KDDLab, University of Pisa and ISTI-CNR, Pisa, Italy\\
\email{\{guidotti,ruggieri\}@di.unipi.it} 
}

\maketitle

\begin{abstract}
Interpretable classification models are built with the purpose of providing a comprehensible description of the decision logic to an external oversight agent. 
When considered in isolation, a decision tree, a set of classification rules, or a linear model, are widely recognized as human-interpretable. 
However, such models are generated as part of a larger analytical process.
Bias in data collection and preparation, or in model's construction may severely affect the accountability of the design process. 
We conduct an experimental study of the stability of interpretable models with respect to feature selection, instance selection, and model selection. Our conclusions should raise awareness and attention of the scientific community on the need of a \textit{stability impact assessment} of interpretable models.
\end{abstract}


\section{Introduction}
\label{sec:introduction}

Interpretable machine learning models aim at trading-off predictive accuracy with human-comprehensibility and verifiability. 
They are also used to explain the global logic of inscrutable black-box machine learning models, such as neural networks, or the local logic of specific decisions taken by such black-boxes~\cite{guidotti2018survey,pedreschi2019meaningful}. 
This is achieved by a form of reverse engineering, where interpretable models are trained on a 
sample of the population or on a random sample in the neighborhood of the instance whose decision has to be explained. 
The data science process of learning an interpretable model, either directly  for decision-making or by reverse engineering a black-box, includes a number of design choices.
\begin{itemize}
\item On the set of features (\textit{feature selection}): a black-box uses a set of features which may be not completely known, hence reverse engineering it must consider which features to use for the surrogate model. 
\item On the subset of instances (\textit{instance selection}): instance generation/perturbation in black-box explanation can be purely random~\cite{ribeiro2016should,haufe2014interpretation}, 
or adopt refined approaches, e.g., the genetic generation process described in~\cite{guidotti2018local}. 
\item On the interpretable model (\textit{model selection}): a growing number of variants and specific learning algorithms are being proposed~\cite{guidotti2018survey}.
\end{itemize}
Such a process must be accountable~\cite{kroll165accountable}, namely the interpretable (possibly, surrogate) model must be able to provide ``a satisfactory answer [about black-box decisions] to an external oversight agent''\footnote{IEEE Glossary for Discussion of Ethics of Autonomous and Intelligent Systems: \textit{https://ethicsinaction.ieee.org}{https://ethicsinaction.ieee.org}.}.
However, since the above design choices include a number of elements subject to randomness, it  may end up with unstable results, i.e., variations in training data collection and selection, or apparently neutral design choices may lead to different interpretable models and decision explanations. Stability of interpretable models is then a key property towards accountability of machine learning (black-box) decision making. 

The contribution of this paper consists of an \textit{experimental study} of the stability of interpretable classification models with respect to the three design choices above.
We will consider \textit{decision trees}, \textit{rule-based classifiers}, and \textit{linear models}, which are widely agreed  to provide explanations of their decisions that are easily interpretable by humans~\cite{freitas2014comprehensible,huysmans2011empirical}.
We conclude that, 
to pursue  accountability, interpretable model's learning processes should comprise a \textit{stability impact assessment} which is currently missing in guidelines and best-practices\footnote{See, e.g., the AINow report 2018 on \textit{Algorithmic impact assessments}: \textit{https://ainowinstitute.org/aiareport2018.pdf}. 
Stability impact assessment differs from sensitivity analysis as the the first one evaluates how much vary the model when different features or records are used to train it, while the last one estimates the impact of varying certain features on the final outcome.}.

The rest of the paper is organized as follows. 
In Section~\ref{sec:related}, we survey related work on the impact of training data variations. 
Section~\ref{sec:background} introduces interpretable models and measures, and feature/instance selection algorithms that will be considered in experiments. 
The evaluation framework is motivated and formalized in Section~\ref{sec:evaluation}. 
Section~\ref{sec:experiments} reports and discusses experimental results. 
Finally, Section~\ref{sec:conclusion} summarizes the contribution of the paper, its limitations, and future work.

\begin{figure*}[t]
\centering
\begin{minipage}[b]{.32\textwidth}
\includegraphics[trim = 0mm 0mm 0mm 0mm, clip,width=\linewidth]{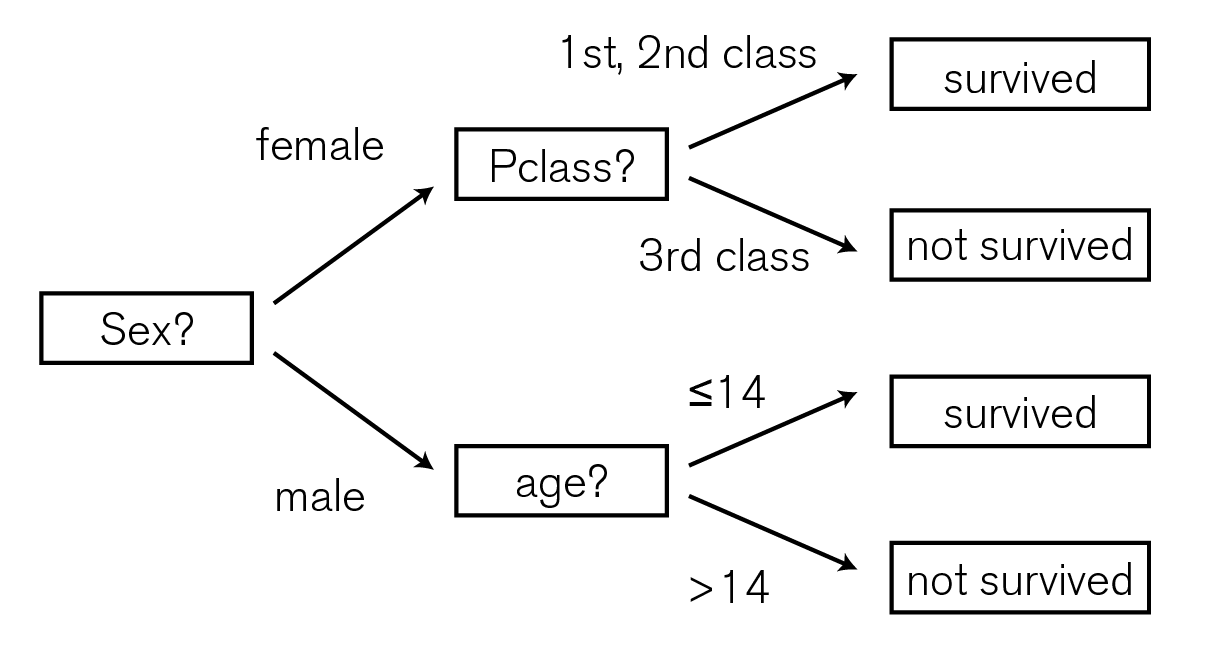}\\
\centering{(a) decision tree}
\end{minipage}%
\begin{minipage}[b]{.25\textwidth}
\begin{lstlisting}
survived <- (age $\leq$ 8) $\wedge$
            (sex = female)
survived <- (age $\geq$ 47) $\wedge$
            (sex = male) $\wedge$
            (Pclass = 1st)
not survived <- (sex = male) $\wedge$
                (Pclass = 3rd)
\end{lstlisting}
\centering{(b) rule-based classifier}
\end{minipage}%
\begin{minipage}[b]{.38\textwidth}
\includegraphics[trim = 10mm 0mm 10mm 0mm, clip,width=\linewidth]{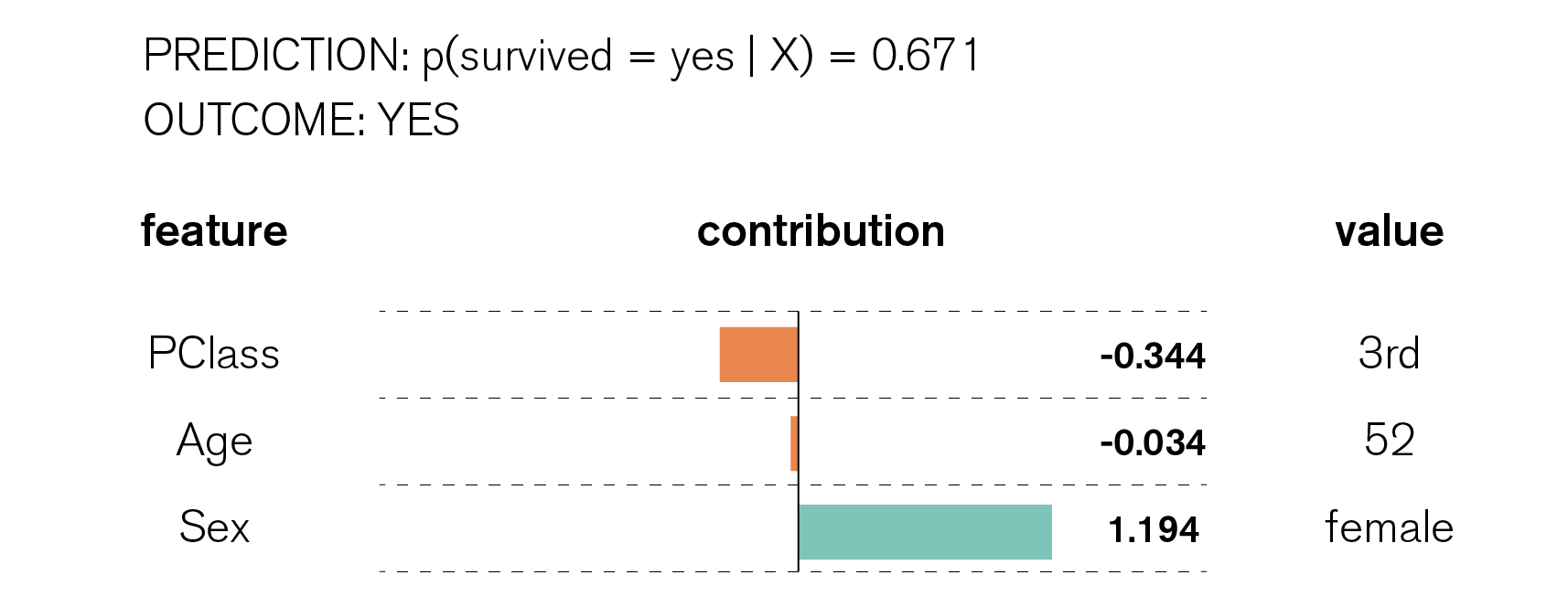}\\
\centering{(c) linear model}
\end{minipage}
\caption{Interpretable models learned on the Titanic dataset
\textit{https://www.kaggle.com/c/titanic}
}
\label{fig:dt}\label{fig:rb}\label{fig:lm}
\vspace{-3mm}
\end{figure*}

\section{Related Work}
\label{sec:related}

Stability is a property of the output of a learning process. The representation of the output can be  \textit{intensional} (a classifier) or \textit{extensional} (its predictions). 

\textit{Extensional stability} of classifier predictions was modelled by~\cite{turney1995bias} through a measure of agreement among predictions. 
The proposed approach consists of a $m \times 2$-fold cross-validation. 
At each of the $m$ steps, two classifiers are built on the two folds, and tested on artificially generated instances from a population distribution. 
The agreement measure is the percentage of instances whose predictions of the two classifiers coincide. 
The average agreement over the $m$ runs is the final estimate of stability of the learning process. 
Agreement is a semantic measure,
and it has the advantage of being classifier-agnostic. Related to measurement of extensional stability is the bias-variance decomposition of the error of classifiers~\cite{trevor2009elements}.  
Bias is reduced and variance is increased with increasing model complexity at the risk of overfitting.
This would suggest that less interpretable models are also more unstable and overfitted. On the theoretical side,~\cite{bousquet2002stability} proved that generalization error can be bound by (expectation of) stability.

Measures of interpretability of classifiers must, however, be necessarily syntactic, since this is the level at which humans interface with models.
This paper concentrates then on \textit{intensional stability} of a learning process. 
One of the early studies regards the impact of training set size on the accuracy of decision trees~\cite{jensen1999effects}, showing that the best performance can be achieved with sufficiently many instances, after which there is no convenience to add more data. Coping with variability of classifiers due to random noise has been tackled by adopting statistical tests for validating split tests at decision nodes~\cite{katz2014confdtree}, or by adopting split methods that account for almost equal split attributes (sources of instability)~\cite{li2002instability}.
Finally, decision tree simplification is another class of approaches that trade-off accuracy with simplicity~\cite{breslow1997simplifying}.
Intensional stability of feature selection methods considered variability in the set of features selected~\cite{kalousis2007stability,nogueira2016measuring}.
Measures of stability include average Jaccard similarity and Pearson's correlation  among all pairs of feature subsets selected from different training sets generated using cross-validation, jacknife or bootstrap. As pointed out by~\cite{kalousis2007stability}, intensional instability of feature selection does not necessarily implies extensional instability of the final classifier, due to redundant features. 
In summary, an experimental study of the intensional instability of interpretable models at the variation of the learning process design choices is missing in the literature. 
This is becoming relevant in the context of black-box explanation, where an early attempt at studying robustness of single explanations is~\cite{alvarez2018robustness,melis2018towards}.

\section{Setting the Stage}
\label{sec:background}

\subsection{Interpretable Models}
Interpretability is the ability to explain or to provide meaning in terms understandable to a human~\cite{guidotti2018survey}. 
Decision trees, rule-based classifiers, and linear models are acknowledged as being interpretable classification models. 

Decision trees (DT) consist of a tree graph with internal nodes representing tests on predictive features, and leaf nodes assigning a class label to instances reaching the leaf (see e.g.,~Fig.~\ref{fig:dt}(a)).
A path from the root to a leaf represents an explanation of the decision at the leaf in terms of a conjunction of test conditions. 
We consider the two mostly adopted greedy learning algorithms: \textbf{CART} (Classification and Regression Trees)~\cite{breiman1984classification} as implemented by the \textit{scikit-learn} Python library\footnote{
\textit{http://scikit-learn.org}.}, and \textbf{C4.5}~\cite{quinlan1993c4} as implemented by the computationally efficient \textit{YaDT} system\footnote{
\textit{http://pages.di.unipi.it/ruggieri/software}.}~\cite{ruggieri2004yadt}. 
\textbf{C4.5} performs multi-way univariate splits and it includes tree simplification (error-based pruning). We do not consider instead the split condition of~\cite{li2002instability}, designed for stability, since it produces disjunctive test conditions, thus leading to a higher expressivity language.

Rule-Based (RB) classifiers consist of a set of classification rules, typically in the form of \textit{if-then} rules stating the class label for a given conjunctive condition on the predictive feature values (see Fig.~\ref{fig:rb}(b)). 
In this work, we consider the \textbf{FOIL} (First Order Inductive Learner)~\cite{quinlan1993foil} and \textbf{CPAR} (Classification based on Predictive Association Rules)~\cite{yin2003cpar}  algorithms, as implemented by the \textit{LUCS-KDD} library\footnote{
\textit{https://cgi.csc.liv.ac.uk/$\sim$frans/KDD/Software}.}. The former generates a very small number of rules, but has lower accuracy than the latter. 
Similarly to DTs, we restrict to sets of conjunctive classification rules. Another natural choice would have been \textbf{RIPPER}~\cite{cohen1995fast}, which however produces \textit{ordered} sequences of conjunctive rules. 
Instead,~we compare DT and RB classifiers with the same expressivity.

Linear Models (LM) classifiers consist of the sign and the magnitude of the contribution of feature values (or ranges) 
to a class label (encoded as an integer) as stated by coefficients in a linear formula (see Fig.~\ref{fig:lm}(c)).
If the contribution is positive (resp.,~negative), the value of the feature increases (resp.,~decreases) the probability of the model's decision.
We focus on three algorithms for linear models: Linear Regression (\textbf{LINREG}) \cite{yan2009linear}, and its regularized forms \textbf{RIDGE}~\cite{tikhonov1963solution} and \textbf{LASSO}~\cite{tibshirani1996regression}, as implemented by the \textit{scikit-learn} library.
They are commonly used in \mbox{black-box explanation approaches~\cite{kononenko2010efficient,ribeiro2016should}.}

\subsection{Measuring Interpretability and Stability}
Several syntactic measures of interpretability are considered in the literature. 
Structural measures (SM) look at models in isolation, and quantify the degree of syntactic (intensional) interpretability of a model by resorting to model complexity. Stability is quantified through the deviation of the measure distribution over models learned from different samples of the population.
Comparative measures (CM) look at pairs of models, and quantify the syntactic  similarity between the two models. Stability is quantified by the mean value over all pairs of models learned from different samples of the population.

Measures common to decision trees, rule-based classifiers and linear models include:
\begin{itemize}
\item \textit{number of features} (SM) used\footnote{While \textit{YaDT} and \textit{LUCS-KDD} work directly on discrete features, algorithms of the \textit{scikit-learn} require binarization of such features. 
Nevertheless, we count the number of original features, not of the binarized ones.}: for DT the features used in at least one split node, for RB those used in at least one rule, for LM the features with non-zero coefficient;
\item \textit{Jaccard coefficient} (CM): the ratio of the number of shared features of two models over the total number of features used by at least one such models.
\item sample \textit{Pearson's} (CM) correlation coefficient~\cite{nogueira2016measuring}: the Pearson's coefficient over the 0/1 vector of features used by two models.
\end{itemize}
Measures specific of a model type include:
\begin{itemize}
    \item for decision trees: \textit{number of nodes} (SM).
    \item for rule-based classifiers: \textit{number of rules} (SM) and \textit{size of rules} (SM), namely the total number of conjuncts in the \textit{if}-part of rules.
    \item for linear models: \textit{Kendall's $\tau$} (CM) rank correlation of coefficients.
\end{itemize}
In summary, for structural measures, one aims at low mean values  (interpretability) and low deviation (stability). For comparative measures, one aims at high mean values (stability) and low deviation (extreme outlier models).

Finally, in order to investigate the relationship between model stability, prediction accuracy, and overfitting, 
we will also compute the F1-scores of models on the training set ($F1_{\mathit{train}}$) and on the test set ($F1_{\mathit{test}}$), and their relative difference
($(F1_{\mathit{train}} - F1_{\mathit{test}})/F1_{\mathit{train}}$), which represents a measure of overfitting.

\subsection{Feature and Instance Selection}
Feature selection (FS)~\cite{Guyon2006} and instance selection (IS)~\cite{olvera2010review} are beneficial in removing noise and redundancies, in reducing the data collection effort, in balancing the data distribution, in speeding up model learning. 
They are supposed to enhance model interpretability by reducing the number of features and by preventing overfitting. 
Both techniques are widely used in reverse engineering of black-box models.
In this paper, we consider the following standard methods, as provided by the 
\textit{scikit-learn}\footnote{
\textit{http://scikit-learn.org/stable/modules/feature\_selection}.} library for feature selection:
\begin{itemize}
\item \textbf{RFE} (Recursive Feature Elimination): given an external estimator that assigns weights to features (a decision tree by default), it greedily removes the least important feature until a given number of features is left (default: half of the total number of features);
\item \textbf{SKB} (Select K Best) removes all but the $k$ top scoring features according to the ANOVA F-value function of the features (default: $K{=}10$);
\item \textbf{SP} (Select Percentile) removes all but a user-specified top scoring percentage of features with respect to the ANOVA F-value (default: $\mathit{pct}{=}10$).
\end{itemize}
Finally, we consider the following instance selection 
methods, as provided by the 
\textit{imba\-lanced-learn}\footnote{
\textit{http://contrib.scikit-learn.org/imbalanced-learn}.} library:
\begin{itemize}
\item \textbf{RUS} (Random Under Sampling) under-samples the majority class by randomly picking instances of the other classes;
\item \textbf{ROS} (Random Over Sampling) over-samples the minority class by replicating instances of that class at random with replacement;
\item \textbf{SMOTE} (Synthetic Minority Over-sampling Technique) 
over-samples minority class by generating instances along the linear segment between an instance of the minority class and one of its $k$ nearest neighbors (default: $k{=}5$). 
\end{itemize}
We restrict here to class balancing and random sampling methods, because they are widely adopted in black-box explanation approaches~\cite{craven1994using,guidotti2018local}. 

\begin{algorithm2e}[t]
	\scriptsize
	\caption{$\mathit{EvaluateStability}(M, X, y)$}
	\label{alg:eval}
	\SetKwInOut{Input}{Input}
	\SetKwInOut{Output}{Output}
    \SetKwInOut{Variables}{Variables}
	\Input{$M$ - classification model, $X$ - dataset,
    $y$ - outcome}
	\Output{$\mathcal{E}$ - evaluations} 
    \Variables{\textit{SM} - structural measures,
    \textit{CM} - comparative measures, \\
    $P$ - pre-processing methods
    }
	\BlankLine
    $\mathcal{M} \leftarrow \emptyset$; $\mathcal{E} \leftarrow \emptyset$ \hfill\texttt{\scriptsize// trained models and  evaluations}\\
    \BlankLine
    \For{$i \in \{1, \dots, 5\}$}{
    	$F \leftarrow \mathit{stratified10Fold}(X, y)$ \hfill\texttt{\scriptsize// 10 fold partitioning}\\
    	\For{$k \in \{1, \dots, 10\}$}{
        	$X', y' \leftarrow F_{-k}(X, y)$ \hfill\texttt{\scriptsize// remove k-th fold}\\
        	$\widehat{X}', \widehat{y}' \leftarrow F_k(X, y)$ \hfill\texttt{\scriptsize// select k-th fold}\\
        	\For{$p \in P$}{
            	$X'', y'' \leftarrow p(X', y')$ \hfill\texttt{\scriptsize// pre-processing}\\
              $m^p_{i, k} \leftarrow \mathit{fit(M, X'', y'')}$ \hfill\texttt{\scriptsize// learn model}\\
              $\mathcal{M} \leftarrow \mathcal{M} \cup \{m^p_{i, k}\}$ \hfill\texttt{\scriptsize// store the model}\\
              $y^* \leftarrow \mathit{predict(m^p_{i, k}, X'')}$ \hfill\texttt{\scriptsize// predict training}\\
              $\widehat{y}^* \leftarrow \mathit{predict(m^p_{i, k}, \widehat{X}')}$ \hfill\texttt{\scriptsize// predict test}\\
              $f^p_{i, k} \leftarrow f1(\widehat{y}',\widehat{y}^*)$\hfill\texttt{\scriptsize// performance}\\
              $\mathcal{P} \leftarrow \mathcal{P} \cup \{f^p_{i, k}\}$\\
              $o^p_{i, k} \leftarrow \frac{f1(y'',y^*)-f1(\widehat{y}',\widehat{y}^*)}{f1(y'',y^*)}$\hfill\texttt{\scriptsize// overfitting}\\
              $\mathcal{O} \leftarrow \mathcal{O} \cup \{o^p_{i, k}\}$\\    }
        }
    }
    \BlankLine
            \For{$p \in P$}{
            $\mathcal{E} \leftarrow \mathcal{E} \cup \{ \underset{m^p_{i, k} \in \mathcal{M}}{avg} f^p_{i, k}\}$ \hfill\texttt{\scriptsize// aggr. performance}\\
            $\mathcal{E} \leftarrow \mathcal{E} \cup \{ \underset{f^p_{i, k} \in \mathcal{M}}{avg} o^p_{i, k}\}$ \hfill\texttt{\scriptsize// aggr. overfitting}\\
            \For{$s \in$ SM}{
            $\mathcal{E} \leftarrow \mathcal{E} \cup \{ \underset{m^p_{i, k} \in \mathcal{M}}{avg} s(m^p_{i, k})\}$ \hfill\texttt{\scriptsize// aggr. $s$}
        } 
    	\For{$c \in$ CM}{
        	$\mathcal{E} \leftarrow \mathcal{E} \cup \{ \underset{m^p_{i, k} \neq \hat{m}^p_{i, k}\in \mathcal{M}}{avg} c(m^p_{i, k}, \hat{m}^p_{i, k}) \}$ \hfill\texttt{\scriptsize// aggr. $c$}
        } 
        }
       \BlankLine
	\Return{$\mathcal{E}$}
\end{algorithm2e}

\begin{table*}[t]
\caption{Experimental datasets.}
\label{tab:datasets}
\small 
\centering
\begin{tabular}{@{}c|cccccccccc@{}}
\toprule
\textbf{dataset} & \textbf{adult} & \textbf{anneal} & \textbf{census} & \textbf{clean1} & \textbf{clean2} & \textbf{coil} & \textbf{cover} & \textbf{credit} & \textbf{sonar} & \textbf{soybean} \\ \midrule
instances & 48,842 & 898 & 299,285 & 476 & 6,598 & 9,822 & 581,012 & 1,000 & 208 & 683 \\
features & 14 & 38 & 40 & 166 & 166 & 85 & 54 & 20 & 60 & 35 \\ 
class values & 2 & 6 & 2 & 2 & 2  & 2 & 7 & 2 & 2 & 19 \\ \bottomrule
\end{tabular}
\end{table*}

\section{Experimental Framework}
\label{sec:experiments}
Interpretable classification models are the end products of an articulated analytic process. 
We will evaluate the impact of process design on their intensional stability. 
To this end, we consider the following steps, which motivate the experimental procedure of Algorithm~\ref{alg:eval}.

First, any observational research project must account for variability/bias in data collection~\cite{danks2017algorithmic}. 
Following a standard methodology for estimating accuracy of classifiers\footnote{Cross-validation is an nearly unbiased estimator of accuracy~\cite{kohavi1995study}. Variability of the estimator is accounted for by adopting repetitions~\cite{kim2009estimating}.}, we adopt a 5-repetition of 10-fold stratified cross-validation to account for variations in the data. 
At each iteration, all the available data is split in 10 folds. 
For each fold, the process described next is applied on 9 folds used as training data, and one fold as test (denoted by the hat $\widehat{\cdot}$).
This is formalized in the two outer loops at lines 2--16 of Algorithm~\ref{alg:eval}. 
Second, the impact of pre-processing steps is evaluated by considering no preprocessing, feature selection, instance selection, and possibly combinations of them. Let $P$ be the a set of pre-processing methods, including no modification at all. The inner loop at lines 7--16 of Algorithm~\ref{alg:eval} iterates over $P$ for the current fold $k$ at iteration $i$. A pre-processing $p \in P$ is applied to the training data, and then the model is learned from the processed data. In Algorithm~\ref{alg:eval}, models are stored in the set $\mathcal{M}$. 
Since pre-processing is the inner loop, paired comparison of models built from different pre-processing methods can be conducted when analyzing the results of the evaluation.
Moreover, lines 13--16  keep track of the predictive performance and of the degree of overfitting on the test data (the $k^{\mathit{th}}$ fold).

Third, measures of interpretability, performance, and overfitting of the learned interpretable classification models must be aggregated over the 50 models (5 repetitions, 10 models each) of each pre-processing method.
Performance, overfitting, and structural measures (SM) are aggregated using the mean value (lines 18--21). 
Comparative measures (CM) are aggregated by taking the all-pairs average (lines 22--23). 
Both loops are inside the loop at lines 17--23 that iterates over the set $P$ of pre-processing algorithms.

The results of the above experimental framework are intended to support  accountability questions that a data analyst should answer before deploying a classification model, namely, how sensitive is the interpretability of a classification model to changes: \textit{in feature selection? in instance selection? in model selection? And, questions such as: which   pre-processing provide the lowest variability for a given model? at which price (e.g.,~accuracy loss)?}

Finally, it is worth noting that the experimental framework does not contain original parts \textit{per se}. All in all, it consists of a repeated stratified cross-validation process plus statistical tests of collected measures of interpretability, stability, accuracy, and overfitting. However, it allows for assessing in a principled way the stability of combinations of old and new feature, instance, and model selection methods. 

\begin{figure}[t]
\includegraphics[trim=2mm 0mm 2mm 0mm, clip, width=0.33\linewidth]{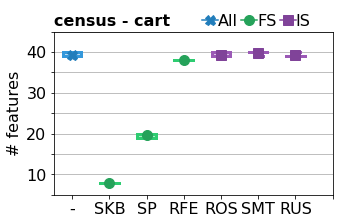}%
\includegraphics[trim=2mm 0mm 2mm 0mm, clip, width=0.33\linewidth]{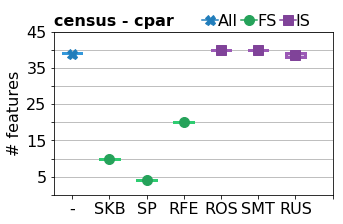}%
\includegraphics[trim=2mm 0mm 2mm 0mm, clip, width=0.33\linewidth]{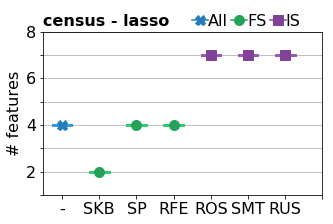}\\
\includegraphics[trim=2mm 0mm 2mm 0mm, clip, width=0.33\linewidth]{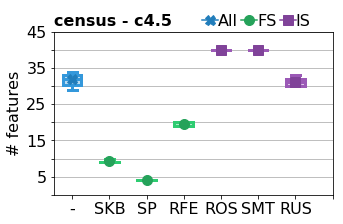}%
\includegraphics[trim=2mm 0mm 2mm 0mm, clip, width=0.33\linewidth]{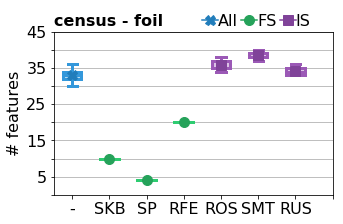}%
\includegraphics[trim=2mm 0mm 2mm 0mm, clip, width=0.33\linewidth]{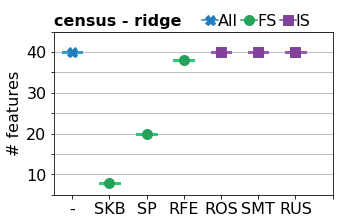}
\caption{Dataset: \textit{census}. Measure: number of features. 
}
\label{fig:census_features}
\end{figure}

\begin{figure}[t]
\includegraphics[trim=2mm 0mm 2mm 0mm, clip, width=0.33\linewidth]{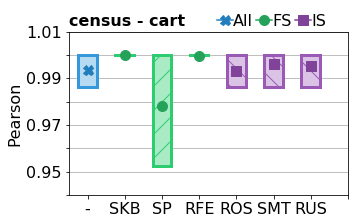}%
\includegraphics[trim=2mm 0mm 2mm 0mm, clip, width=0.33\linewidth]{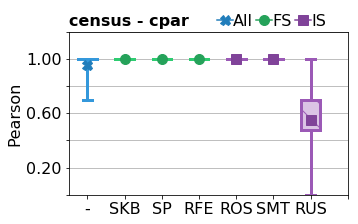}%
\includegraphics[trim=2mm 0mm 2mm 0mm, clip, width=0.33\linewidth]{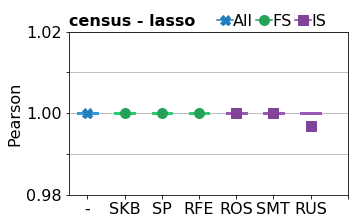}\\[2mm]
\includegraphics[trim=2mm 0mm 2mm 0mm, clip, width=0.33\linewidth]{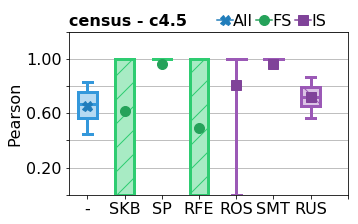}%
\includegraphics[trim=2mm 0mm 2mm 0mm, clip, width=0.33\linewidth]{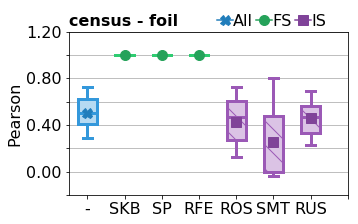}%
\includegraphics[trim=2mm 0mm 2mm 0mm, clip, width=0.33\linewidth]{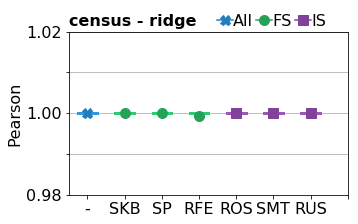}
\caption{Dataset: \textit{census}. Measure: Pearson's correlation. 
}
\label{fig:census_pearson}
\vspace{-5mm}
\end{figure}

\begin{table*}[t]\scriptsize
	\caption{Number of features, Pearson's correlation, and F1-score (mean $\pm$ stdev).}
	\centering
	\scalebox{0.95}{\noindent
		\begin{tabular}{c|rrr|rrr|rrr}
        \toprule
        {} & \multicolumn{3}{c|}{\textbf{number of features}} & \multicolumn{3}{c|}{\textbf{Pearson's correlation}} & \multicolumn{3}{c}{\textbf{F1-score}} \\
        \textbf{dataset} & \multicolumn{1}{c}{\textbf{C4.5}} & \multicolumn{1}{c}{\textbf{RFE+C4.5}} & \multicolumn{1}{c|}{\textbf{C4.5 d $\leq$ 5}} & \multicolumn{1}{c}{\textbf{C4.5}} & \multicolumn{1}{c}{\textbf{RFE+C4.5}} & \multicolumn{1}{c|}{\textbf{C4.5 d $\leq$ 5}} & \multicolumn{1}{c}{\textbf{C4.5}} & \multicolumn{1}{c}{\textbf{RFE+C4.5}} & \multicolumn{1}{c}{\textbf{C4.5 d $\leq$ 5}}  \\
        \midrule
        \textbf{adult} & 14.00 $\pm$ 0.00 & 7.00 $\pm$ 0.00 & \textbf{6.24 $\pm$ 1.09} & \textbf{1.00 $\pm$ 0.00} & \textbf{1.00 $\pm$ 0.00} & 0.76 $\pm$ 0.19 & \textbf{0.85 $\pm$ 0.00} & \textbf{0.85 $\pm$ 0.00} & 0.84 $\pm$ 0.00 \\
        \textbf{anneal} & 11.20 $\pm$ 0.69 & 9.04 $\pm$ 1.06 & \textbf{5.76 $\pm$ 0.43} & 0.85 $\pm$ 0.09 & 0.66 $\pm$ 0.17 & \textbf{0.88 $\pm$ 0.11} & \textbf{0.98 $\pm$ 0.01} & \textbf{0.98 $\pm$ 0.01} & 0.85 $\pm$ 0.03 \\
        \textbf{census} & 31.92 $\pm$ 1.92 & 19.66 $\pm$ 0.47 & \textbf{7.78 $\pm$ 0.70} & 0.66 $\pm$ 0.15 & 0.49 $\pm$ 0.50 & \textbf{0.89 $\pm$ 0.07} & \textbf{0.95 $\pm$ 0.00} & \textbf{0.95 $\pm$ 0.00} & 0.93 $\pm$ 0.00 \\
        \textbf{clean1} & 23.70 $\pm$ 2.82 & 24.68 $\pm$ 2.73 & \textbf{6.30 $\pm$ 2.16} & 0.25 $\pm$ 0.15 & 0.03 $\pm$ 0.13 & \textbf{0.48 $\pm$ 0.25} & \textbf{0.83 $\pm$ 0.05} & 0.81 $\pm$ 0.05 & 0.68 $\pm$ 0.06 \\
        \textbf{clean2} & 74.80 $\pm$ 4.78 & 60.06 $\pm$ 3.64 & \textbf{13.26 $\pm$ 0.77} & 0.37 $\pm$ 0.08 & -0.45 $\pm$ 0.22 & \textbf{0.59 $\pm$ 0.12} & \textbf{0.97 $\pm$ 0.01} & \textbf{0.97 $\pm$ 0.01} & 0.89 $\pm$ 0.01 \\
        \textbf{coil} & 8.12 $\pm$ 3.68 & 5.34 $\pm$ 3.85 & \textbf{3.70 $\pm$ 3.53} & 0.20 $\pm$ 0.18 & 0.18 $\pm$ 0.31 & \textbf{0.27 $\pm$ 0.37} & \textbf{0.91 $\pm$ 0.00} & \textbf{0.91 $\pm$ 0.00} & \textbf{0.91 $\pm$ 0.00} \\
        \textbf{cover} & 52.78 $\pm$ 0.50 & 27.00 $\pm$ 0.00 & \textbf{16.42 $\pm$ 0.70} & 0.71 $\pm$ 0.36 & 1.00 $\pm$ 0.00 & \textbf{0.92 $\pm$ 0.05} & \textbf{0.95 $\pm$ 0.00} & 0.94 $\pm$ 0.00 & 0.45 $\pm$ 0.00 \\
        \textbf{credit} & 17.56 $\pm$ 1.44 & 9.66 $\pm$ 0.51 & \textbf{10.70 $\pm$ 1.76} & 0.12 $\pm$ 0.29 & 0.32 $\pm$ 0.79 & \textbf{0.43 $\pm$ 0.22} & 0.70 $\pm$ 0.04 & \textbf{0.71 $\pm$ 0.04} & \textbf{0.71 $\pm$ 0.04} \\
        \textbf{sonar} & 12.04 $\pm$ 1.44 & 11.50 $\pm$ 1.46 & \textbf{9.32 $\pm$ 1.83} & 0.23 $\pm$ 0.18 & -0.06 $\pm$ 0.22 & \textbf{0.28 $\pm$ 0.21} & \textbf{0.74 $\pm$ 0.08} & 0.73 $\pm$ 0.08 & 0.73 $\pm$ 0.09 \\
        \textbf{soybean} & 22.52 $\pm$ 1.70 & 15.42 $\pm$ 1.17 & \textbf{14.82 $\pm$ 1.71} & 0.70 $\pm$ 0.13 & -0.80 $\pm$ 1.06 & \textbf{0.79 $\pm$ 0.12} & \textbf{0.91 $\pm$ 0.03} & 0.89 $\pm$ 0.03 & 0.68 $\pm$ 0.03 \\
        \bottomrule
        \end{tabular}
    }
	\label{tab:nf:pc}
\end{table*}

\section{Experiments}
\label{sec:experiments}
We run experiments on a selection of ten small and medium sized datasets widely referenced for classification tasks and publicly available from the UCI ML repository\footnote{
\textit{https://archive.ics.uci.edu/ml/datasets.html}}. 
Table~\ref{tab:datasets} shows summary statistics on the datasets: instances are in the range 208--581K, features in 14--166, and number of classes in 2--19.
The framework of Algorithm~\ref{alg:eval} has been implemented in Python\footnote{Source code and datasets: 
\textit{https://github.com/riccotti/InterpretableModels}.}
by integrating external libraries (\textbf{YADT}, \textbf{FOIL}, and \textbf{CPAR}) 
through wrappers of inputs/outputs. 
The software has been designed to be extensible to additional models, pre-processing methods, and intepretability measures. 
Unless specified otherwise, parameters of algorithms are the defaults in their original systems\footnote{\textbf{C4.5}: split = Gain Ratio, stop criterion = -m 2, pruning = -ebp (error-based); \textbf{CART}: split = Gini, min\_samples\_split = 2, min\_samples\_leaf = 1, max\_depth = None; \textbf{CPAR}: delta = 0.05, alpha = 0.3, gain\_similarity\_ratio = 0.99, min\_gain\_thr = 0.7; \textbf{FOIL}: min\_gain\_thr = 0.7; \textbf{LASSO}: alpha = 1.0; \textbf{RIDGE}: alpha = 1.0.
FS: \textbf{SBK}: k=10, fun=ANOVA F-value, \textbf{SP}: percentile=10, fun=ANOVA F-value, \textbf{RFE}: n\_features\_to\_select = half of the features, estimator=sklearn DecisionTree default parameters. IS: \textbf{SMT}: k=5.
}.

\subsection{Preliminary Analysis on the Census Dataset}\label{sec:census}
Let us start focusing on the number of features used by a classification model. Fig.~\ref{fig:census_features} considers the \textit{census} dataset. 
Left plots report on DT models (\textbf{CART} and \textbf{C4.5}), middle plots on RB models (\textbf{CPAR} and \textbf{FOIL}), and right plots on LM models (\textbf{LASSO} and \textbf{RIDGE}\footnote{We omit \textbf{LINREG} for space reasons as it behaves as \textbf{RIDGE}.}). Each plot shows the boxplots for no pre-processing (``-"), for 3 Feature Selection (FS) methods (\textbf{SKB}, \textbf{SP}, and \textbf{RFE}), and for 3 Instance Selection (IS) methods (\textbf{ROS}, \textbf{SMT}, and \textbf{RUS}). Feature selection methods reduce the total number of features used by the classification model, as one would expect, thus improving the interpretability measure. 
Moreover, since redundant/noisy features are removed as well, this also reduces deviation over the 50 folds, thus improving stability. 
Instance selection has a similar beneficial effect on deviation, but in some cases (\textbf{LASSO} and \textbf{C4.5}) it increases the number of features. 
However, for a gross-grained measure such as the number of features, the low variability provides a distorted indication of stability. In fact, two models may still largely differ in the set of features used while the number of such features is the same for both models.
Jaccard similarity or Pearson's correlation among all pairs of feature sets across the 50 folds can better measure variability of the set of features (rather than the number of them) used by a classifier. 
Fig.~\ref{fig:census_pearson} reports Pearson's correlation for the \textit{census} dataset. 
We omit the Jaccard measure for lack of space and because it yields similar patterns. Linear models are stable, independently from the pre-processing method. In fact, Pearson's correlation is always very close to $1$. 
For rule-based models, FS also leads to stable models. 
IS increases deviation of Pearson's correlation for rule-based and decision trees classifiers. 
This means that extreme outlier models (in terms of feature's vector) become more frequent.  

\subsection{Preliminary Analysis on Decision Trees}\label{sec:dt}
It is worth looking into details to the case of  decision trees. Table~\ref{tab:nf:pc} shows the number of features and Pearson's correlation for three \textbf{C4.5} classifiers: the one built on all available features, the one built on features selected by \textbf{RFE}, and the one built on all features but restricted to a maximum depth of 5 (\textbf{d $\leq$ 5}). Setting a maximum depth is a valid approach to enhance the interpretability of decision trees. Such a method performs the best both w.r.t.~the number of features and the Pearson's correlation. This is achieved at the expenses of loss in accuracy, as shown by the F1 score columns in Table~\ref{tab:nf:pc}.
Contrasting \textbf{C4.5} alone with \textbf{RFE+C4.5}, we observe that addition of the \textbf{RFE} method reduces the number of used features (interpretability), at the expenses of Pearson's correlation (stability), while the F1-score (accuracy) remains almost the same.

\begin{figure}[t]
\centering
\includegraphics[trim=2mm 0mm 2mm 0mm, clip, width=\linewidth]{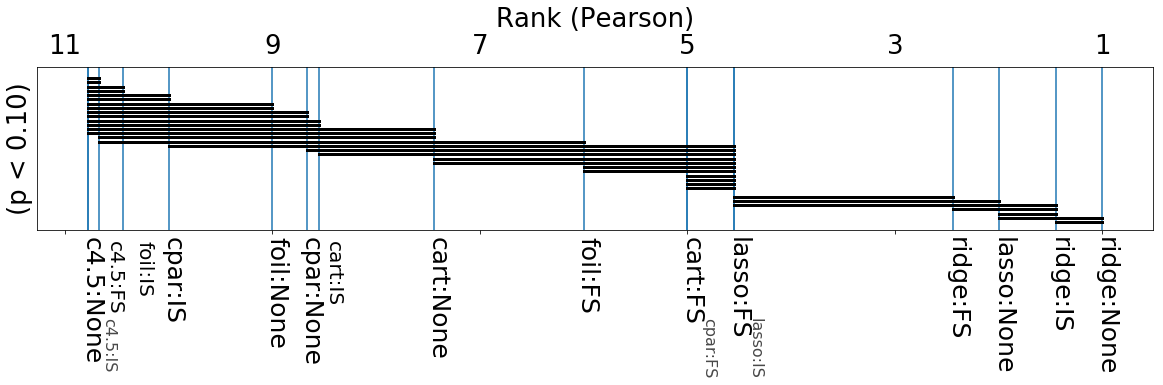}
\caption{Comparison of model's rank w.r.t. Pearson's correlation against each other with the Nemenyi test. Groups of classifiers that are not significantly different at $90$\% significance level are connected. Best ranks on the right.}
\label{fig:wilcoxon_pearson}
\end{figure}

\begin{figure*}[t]
\centering
\includegraphics[trim=2mm 0mm 2mm 0mm, clip, width=0.23\linewidth]{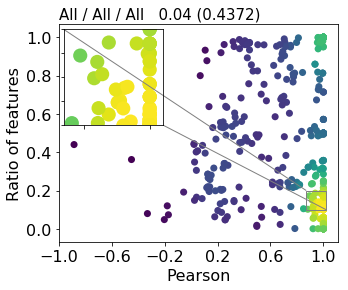}%
\includegraphics[trim=2mm 0mm 2mm 0mm, clip, width=0.23\linewidth]{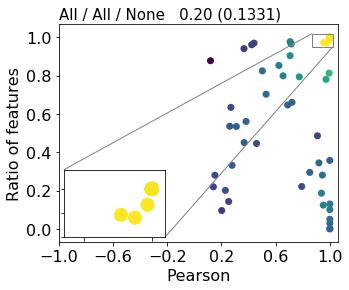}%
\includegraphics[trim=2mm 0mm 2mm 0mm, clip, width=0.23\linewidth]{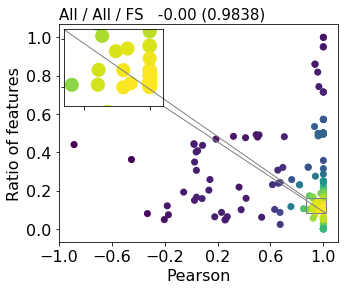}%
\includegraphics[trim=2mm 0mm 2mm 0mm, clip, width=0.23\linewidth]{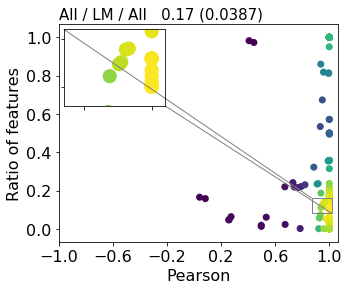}%
\caption{Scatter density plots of Pearson's correlation vs Ratio of features used. Each point's coordinates are the mean values over the 50 experimental folds of some dataset for the following conditions (left to right): experiments for all datasets/classifiers/pre-processing; experiments with no pre-processing; experiments with FS;  experiments with LM. On top: correlation and p-value. Colors: yellow = high density, green = medium density, purple = low density.
}
\label{fig:scatter1}
\end{figure*}

\begin{figure*}[tb]
\centering
\includegraphics[trim=2mm 0mm 2mm 0mm, clip, width=0.23\linewidth]{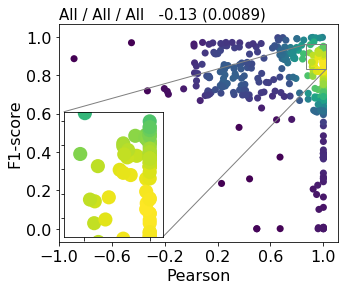}%
\includegraphics[trim=2mm 0mm 2mm 0mm, clip, width=0.23\linewidth]{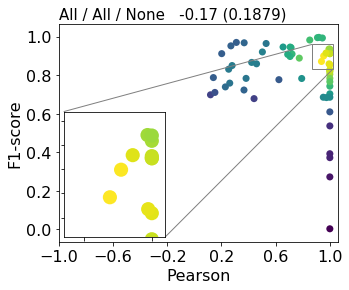}%
\includegraphics[trim=2mm 0mm 2mm 0mm, clip, width=0.23\linewidth]{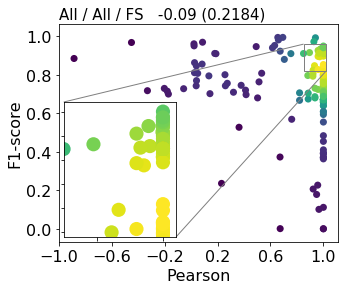}%
\includegraphics[trim=2mm 0mm 2mm 0mm, clip, width=0.23\linewidth]{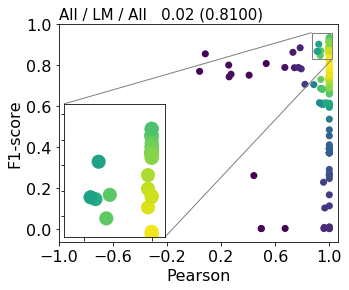}%
\caption{Scatter plots of Pearson vs F1-score. See caption of Fig.~\ref{fig:scatter1}.}
\label{fig:scatter}
\vspace{-3mm}
\end{figure*}

\subsection{Statistical Comparison of Models' Stability}
The previous two subsections demonstrate how the exploration of the experimental results can provide insights when considering a specific dataset (Section~\ref{sec:census}) or a specific model (Section~\ref{sec:dt}). The rest of the experiments will be devoted, instead, to understand whether there are general patterns among all possible usages of IS, FS, and models.

First, we compare the various methods among them with reference to stability.
The non-parametric Friedman test compares the average ranks of learning methods over multiple datasets w.r.t. an evaluation measure, in our case Pearson's correlation. The null hypothesis that all methods are equivalent is rejected ($p < .001$).
The comparison of the ranks of all methods against each other can be visually represented as shown in Fig.~\ref{fig:wilcoxon_pearson} (see~\cite{demvsar2006statistical} for details).
The post-hoc Nemenyi test is used to connect methods that are not significantly different among each other. Linear models have the best ranks. For a fixed classifier, models obtained using feature selection pre-processing rank better than methods without. \textit{Instance selection methods and decision trees} have the lowest ranks, i.e.,~they \textit{are the most unstable with respect to the set of features} used by the learned model.

\subsection{Stability-Interpretability}
Let us now investigate patterns of correlation between interpretability and stability through the scatter density plots in Fig.~\ref{fig:scatter1}, where Pearson's index (stability) is plotted against the ratio of the number of used features over the total number of features (interpretability).
There are 4 scatter plots. Each point represents an experiment (50 folds).
From left to right and top to bottom: experiments for all datasets/classifiers/\-pre-processing, experiments for all datasets and classifiers but only those with no pre-processing, experiments for all datasets and classifiers but only those with feature selection pre-processing, and experiments for only linear model classifiers. 
Numbers on top of scatter plots are linear correlation and, in parenthesis, p-values of such correlation. The top left plot does not highlight correlation between the measures of stability and interpretability, in general. 
Using no pre-processing methods increase the correlation (higher stability means lower interpretability). 
Feature selection does not impact on the correlation. 
Finally, the rightmost plot shows some positive correlation for linear models at 95\% significance level.

\begin{figure}[tb]
\centering
\includegraphics[trim=2mm 0mm 2mm 0mm, clip, width=\linewidth]{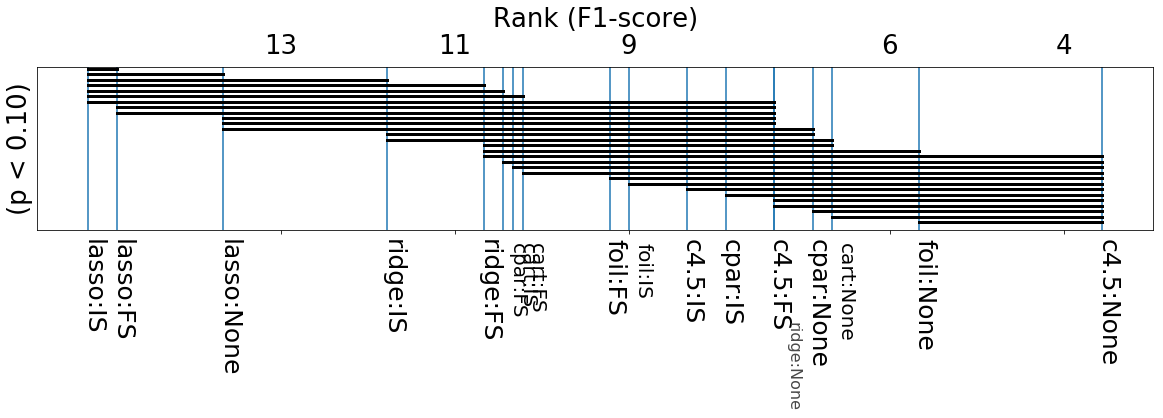}
\vspace{-2mm}
\caption{Same as Fig.~\ref{fig:wilcoxon_pearson} but w.r.t. the F1-score.}
\label{fig:wilcoxon_f1}
\end{figure}

\begin{figure}[tb]
\centering
\includegraphics[trim=2mm 0mm 2mm 0mm, clip, width=0.5\linewidth]{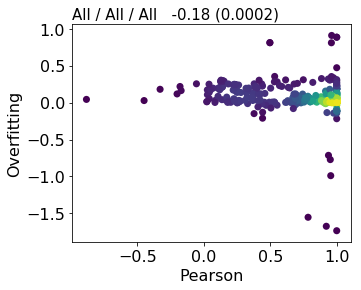}%
\includegraphics[trim=2mm 0mm 2mm 0mm, clip, width=0.5\linewidth]{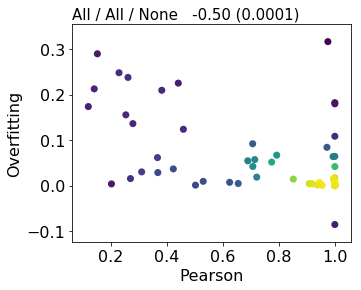}
\caption{Pearson vs Overfitting.
Left: all experiments; right: no pre-processing. On top: correlation and p-value.}
\label{fig:scatter_overfit}
\end{figure}

\subsection{Stability-Accuracy} 
We now investigate the relation between stability and predictive accuracy.
Fig.~\ref{fig:scatter} reports scatter density plots of Pearson's correlation against F1 score (averaged over  the 50 test folds) for the same conditions as in Fig.~\ref{fig:scatter1}.
There is a small negative correlation for the three leftmost plots: the higher the stability (high values of Pearson's correlation) the lower the accuracy (low values of F1-score). Such a correlation is statistically significant at 99\% confidence level only for the leftmost plot, which includes all experiments.
For linear models, the two measures are uncorrelated. Let us look more in detail to the case of all experiments.
Fig.~\ref{fig:wilcoxon_f1} compares the ranks of the various models w.r.t.~the F1 measure averaged over the 50 experimental folds. 
Ranks are approximately symmetric to the ones of Pearson's correlation shown in Fig.~\ref{fig:wilcoxon_pearson}.
Decision trees and rule-based classifiers are the best performing. Linear models are at the bottom of the ranking. 
Instance selection does not improve ranks of classifiers. In summary, for the interpretable models considered here, \textit{stability and accuracy are contrasting objectives}. This demands for a trade-off analysis.

\subsection{Stability-Overfitting}
Fig.~\ref{fig:scatter_overfit} reports scatter plots of stability vs overfitting, defined as the relative difference of F1 accuracy between training and test set averaged over 50 folds. A negative correlation is clearly observed and statistically significant: higher Pearson's correlation (stability) leads to smaller overfitting (generalizability). This is more apparent in experiments with no pre-processing (right in Fig.~\ref{fig:scatter_overfit}).
Such a result is somehow expected, due to the bias-variance decomposition mentioned earlier~\cite{trevor2009elements}.
In summary, \textit{stability and overfitting are contrasting objectives}.

\begin{figure}[tb]
\centering
\includegraphics[trim=2mm 0mm 2mm 0mm, clip, width=0.48\linewidth]{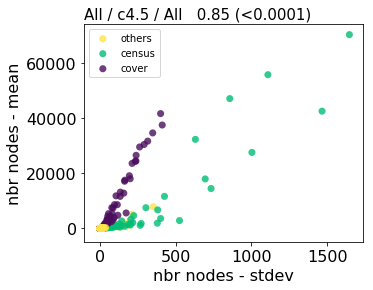}%
\includegraphics[trim=2mm 0mm 2mm 0mm, clip, width=0.48\linewidth]{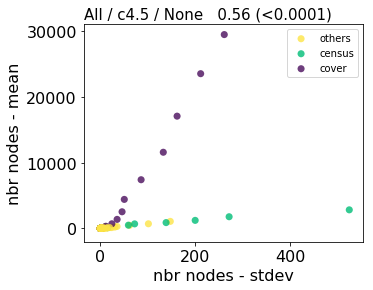}
\caption{Scatter plot of stability (deviation) vs interpretability (mean) w.r.t. number of nodes, left: all \textbf{C4.5} exp.'s; right: \textbf{C4.5} with no pre-processing.}
\label{fig:scatter_tree_color}
\end{figure}

\begin{figure}[tb]
\centering
\includegraphics[trim=2mm 0mm 2mm 0mm, clip, width=0.48\linewidth]{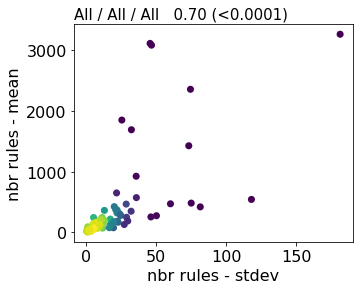}%
\includegraphics[trim=2mm 0mm 2mm 0mm, clip, width=0.48\linewidth]{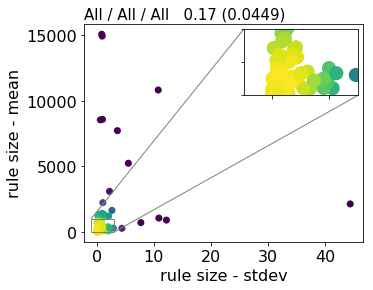}
\caption{Scatter density plot of stability (stddev) vs interpretability (mean) w.r.t. nbr of rules (left) and size of rules (right) in RB models. All experiments.}
\label{fig:scatter_rules}
\end{figure}

\begin{figure}[t]
\includegraphics[trim=2mm 0mm 2mm 0mm, clip, width=0.34\linewidth]{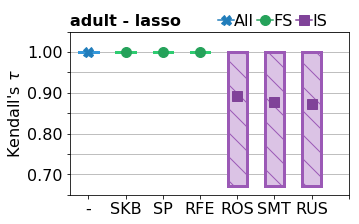}%
 \includegraphics[trim=2mm 0mm 2mm 0mm, clip, width=0.34\linewidth]{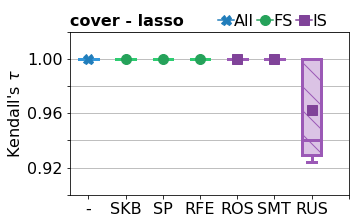}%
 \includegraphics[trim=2mm 0mm 2mm 0mm, clip, width=0.34\linewidth]{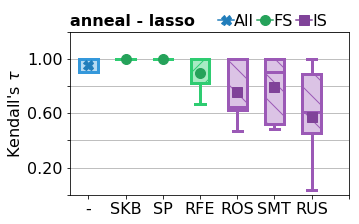}\\
 \includegraphics[trim=2mm 0mm 2mm 0mm, clip, width=0.34\linewidth]{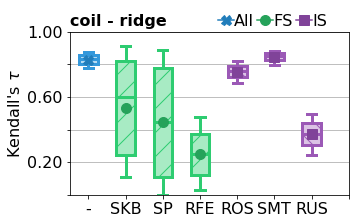}%
  \includegraphics[trim=2mm 0mm 2mm 0mm, clip, width=0.34\linewidth]{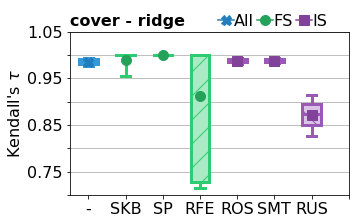}%
  \includegraphics[trim=2mm 0mm 2mm 0mm, clip, width=0.34\linewidth]{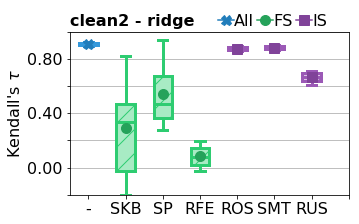}%
 \caption{Measure: Kendall's $\tau$. Models: LM.}
 \label{fig:LM_kendalltau}
 \end{figure}

\subsection{Model-Specific Measures}
When restricting to specific classifiers, finer-grained measures of interpretability can be adopted. Let us start considering the number of nodes in decision trees (for the tree depth measure, we obtain similar findings).  
We study the relation between interpretability and stability by varying the stopping parameter in tree construction from $m{=}2$ (default value) to $m{=}$half of the size of the dataset using a geometric progression. Such parameter stops node splitting during tree construction if the number of cases at the node is below the threshold $m$. Thus, we can control the maximum size of a decision tree. 
Fig.~\ref{fig:scatter_tree_color} shows the scatter plot of mean number of nodes vs standard deviation of the number of nodes over the 50 experimental folds. A statistically significant positive correlation is clearly visible, especially when restricting to a dataset in isolation (experiments with the two largest datasets are shown in different colors).

For rule-based classifiers, Fig.~\ref{fig:scatter_rules} shows the stability-interpretability relation in terms of number of rules (left) and size of rules (right). Each point has coordinates the standard deviation (x-axis) and the mean (y-axis) number/size of rules over the 50 experimental folds. Basically, the two plots are RB-specific versions of the density scatter plots in Fig.~\ref{fig:scatter1}. Contrasting the two figures, there is now a larger statistically significant positive correlation between stability and interpretability. The correlation for the finer grained measure of sizes of rules is smaller than for the gross grained measure of number of rules, which is somehow expected.

Finally, let us consider linear models. Kendall's $\tau$ measures the rank correlation of two sets of features, where the rank of a feature is calculated w.r.t. the descending absolute value of its coefficient. 
Fig.~\ref{fig:LM_kendalltau} reports the boxplots of $\tau$'s  values over the 50 experimental folds. 
\textbf{LASSO} is generally more stable than \textbf{RIDGE} (high values of $\tau$). 
Feature selection increases variability of the measure (extreme outlier models) for \textbf{RIDGE}, but not for \textbf{LASSO}. Vice-versa, IS increases variability for \textbf{LASSO}, but not for \textbf{RIDGE}.

\subsection{Discussion}
Experimental results highlight how the data science process of learning classifiers suffers from some variability in the measures of interpretability of produced models. 
There is a tension between optimizing predictive accuracy from one side, and intensional stability of interpretable classifiers on the other side. Stability and generalizability appear to be common goals, or, stated otherwise, stability and overfitting appear contrasting objectives. Also, stability and interpretability appear to be slightly positively correlated. Existing approaches for improving generalizability of classifiers, however, cannot be always applied to interpretable models. Aggregation methods (e.g., bagging, boosting, random forests) produce models that are widely agreed to be difficult to interpret. 
We claim that the data analyst should conduct a \textit{stability impact assessment} together with predictive performance analysis in order to alleviate the tension between the two objectives. Such a stability impact assessment amounts at analysing the empirical distribution of the relevant interpretability measures at the variation of the design choices. Which measure is the most relevant is another design aspect that must be accounted for and that may affect the stability impact assessment. E.g.,~the two scatter density plots in Fig.~\ref{fig:scatter_rules} show different correlation between interpretability and stability for different measures.
In summary, by exploiting the experimental framework proposed in this paper, a trade-off assessment allows for evaluating the impact of design choices made over a collection of candidate models and pre-processing methods. Overall, it provides evidence that the data analyst has been accountable for one's conduct.

\section{Conclusion}
\label{sec:conclusion}
The intended objective of this paper is to raise awareness by the machine learning community on the issue of being accountable in the design of classification models, particularly those used for socially sensitive decision making. 
Accountability is more than interpretability, and it requires, in our opinion, an impact assessment of the whole learning process. 
The case of interpretable models is challenging; being them comprehensible ``by definition", there is the risk that data analysts overlook the issue of accountability of the extraction process as a whole. In concrete, 
explanations of automated decisions made by black box 
models, such as neural networks, may suffer from data selection/processing bias, which makes it hard argumenting on the decision, e.g.,~in a trial before~a~court.

Our main contributions consist of a framework for intensional stability impact assessment, and an experimental analysis on several pre-processing methods and classification algorithms. The approach is implemented, released as open source, and extensible to new classifiers, methods, and measures. 
Experimental results show that the studied interpretable models exhibit considerable variability in terms of structural and comparative measures. Interpretability of linear models appears to be more stable than for other models, but at the expenses of lower accuracy. Decision trees, on the other hand, exhibit more variability, but they are more accurate. Stability is clearly negatively correlated to accuracy and to overfitting. However, no other generally valid pattern can be drawn. 
Thus, in parallel to optimizing predictive accuracy, practitioners have to look at the impact on intensional stability as well. 

Several extensions of our framework are possible. 
First, for sake of space, we considered only a limited number of interpretable models, pre-processing methods, datasets, and measures. 
E.g.,~the comparative measure of tree edit distance \cite{schwarz2017new} is even more fine-grained than  decision tree size. 
Second, with the exception of Table~\ref{tab:nf:pc} and Fig.~\ref{fig:scatter_tree_color}, we did not consider parameters of the learning algorithms and pre-processing methods. This would add a further loop to Algorithm~\ref{alg:eval}, where parameters are optimized from a parameter space (uniformly, greedly, etc.). 
Third, we considered only objective measures of interpretability and stability. 
A lab experiment can test subjective measures (legibility, understantability) on a pool of actual users. 
Fourth, we concentrated on interpretability of models as a whole, possibly surrogate models of black-boxes. Interpretability of explanations of individual black-box decisions is also worth considering. We could extend  Algorithm~\ref{alg:eval} to collect explanations for the instances in the test fold data, and then to compute measures of interpretability of such explanations.

\section*{Acknowledgment}
Work partially supported by the European Community H2020 programme under the funding schemes: ``INFRAIA-1-2014-2015: Research Infrastructures'' G.A. 654024 \emph{``SoBigData''} 
({
\textit{www.sobigdata.eu}}), 
G.A. 78835 \emph{``Pro-Res''} ({
\textit{prores-project.eu}}), and 
G.A. 825619 \emph{``AI4EU''} ({
\textit{www.ai4eu.eu}
}).

\bibliographystyle{abbrv}
\bibliography{biblio}

\end{document}